\PassOptionsToPackage{table}{xcolor}

\documentclass{bmvc2k}

\usepackage{booktabs}
\usepackage{amsfonts}
\usepackage{amssymb}
\usepackage{nicefrac}
\usepackage{array}
\usepackage{multirow}
\usepackage{enumitem}
\usepackage{float}

\usepackage[capitalize]{cleveref}

\title{QCell: Recombining and Aligning\\
Cell Queries for Overlapping\\
Instance Segmentation}

\newcommand{\methodname}{QCell}
\definecolor{suppred}{RGB}{202, 56, 79}

\addauthor{Yaroslav Prytula}
{yaroslav.prytula@ut.ee}
{1,2}

\makeatletter
\bmv@sbox{authmail1}{%
    \small\textcolor{bmv@sectioncolor}{%
        \shortstack[l]{%
            \bmvaUrl{yaroslav.prytula@ut.ee}\\
            \bmvaUrl{s.prytula@ucu.edu.ua}%
        }%
    }%
}
\makeatother

\addauthor{Anton Popov}
{a.popov@ucu.edu.ua}
{2,3}

\addauthor{Dmytro Fishman}
{dmytro.fishman@ut.ee}
{1,4,5}

\addinstitution{
Institute of Computer Science\\
University of Tartu, Tartu, Estonia
}

\addinstitution{
Faculty of Applied Sciences\\
Ukrainian Catholic University,\\
Lviv, Ukraine
}

\addinstitution{
Department of Electronic Engineering,\\
Micro- and Biomedical Electronics\\
Igor Sikorsky Kyiv Polytechnic Institute,\\
Kyiv, Ukraine
}

\addinstitution{
STACC OÜ, Tartu, Estonia
}

\addinstitution{
Better Medicine OÜ, Tartu, Estonia
}

\runninghead{Prytula, Popov, Fishman}
{QCell: Recombining and Aligning Cell Queries}

\definecolor{our_results_color}{rgb}{0.95,0.95,0.95}
\definecolor{lime}{rgb}{0.2,0.9,0.2}

\newcommand{\smallstd}[1]{%
  \raisebox{-0.35ex}{\scalebox{0.80}{$(\pm#1)$}}%
}

\crefname{section}{Sec.}{Secs.}
\Crefname{section}{Section}{Sections}
\crefname{table}{Tab.}{Tabs.}
\Crefname{table}{Table}{Tables}

\begin{document}

\maketitle



\vspace{-2.5em}
\begin{abstract}
Instance segmentation of overlapping cells in microscopy remains challenging due to semi-transparent structures that produce weak boundaries and mixed visual evidence in overlap regions. Existing methods address this through local regions of interest or shape priors but lack global reasoning across overlapping objects. We present QCell, a novel query-based model that de-overlaps cell instances in microscopy scenes. Our approach combines (i) an instance recombination module that decomposes and recombines query representations in latent space, enabling the model to reason about complete object structure under overlap, and (ii) a contrastive query alignment objective that combines distinctive instance feature learning and separation of overlapping cell queries. We additionally introduce a new Organoid dataset benchmark for overlapping cell segmentation. We show that QCell outperforms state-of-the-art methods across multiple benchmarks, achieving \textit{+2.2 AP} and \textit{+2.7 AJI} on \textit{ISBI2014}. Code is available at \url{https://github.com/SlavkoPrytula/QCell}
\end{abstract}

\vspace{-1.5em}
\begin{center}
    \includegraphics[width=0.80\textwidth]{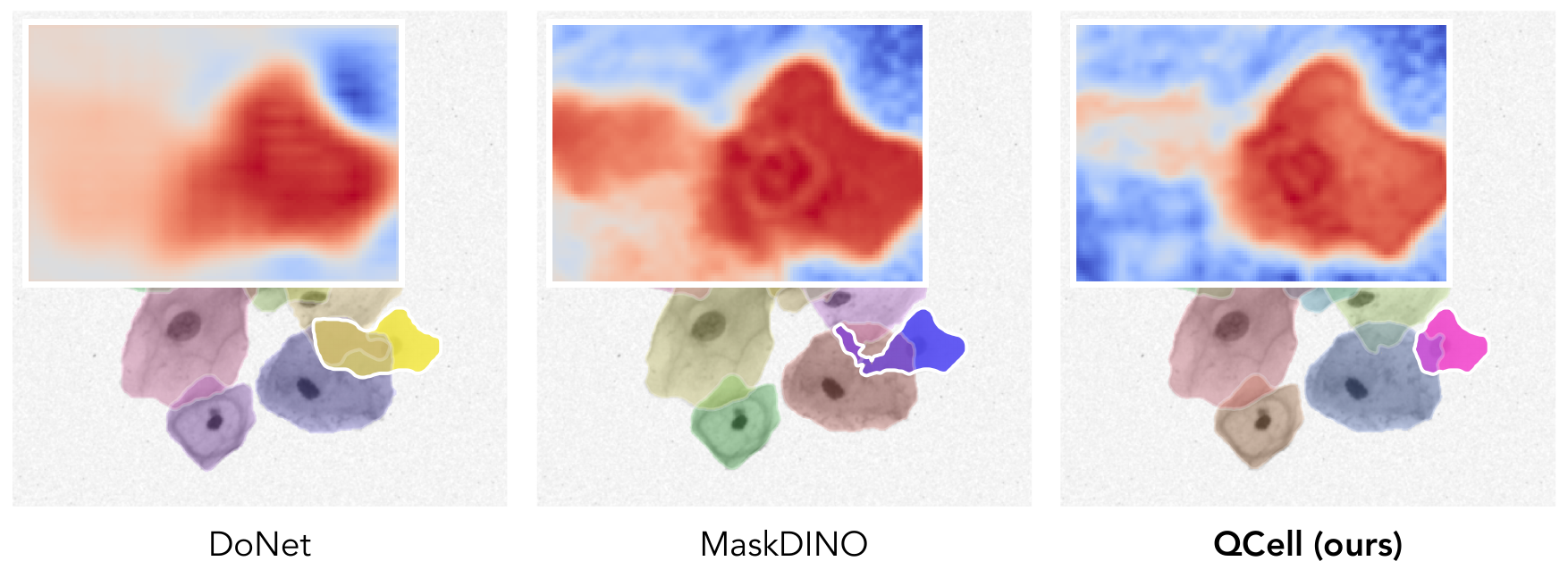}
    \refstepcounter{figure}
    \label{fig:teaser}
    \vspace{-0.3em}

    {\small
    \textbf{Figure~\thefigure.}
    Overlapping cells produce weak boundaries and ambiguous visual evidence. Compared to existing methods, QCell better preserves complete structures and separates neighboring instances.
    }
\end{center}
\section{Introduction}
\label{sec:introduction}


Cell instance segmentation is a fundamental step in microscopy image analysis, enabling downstream measurements of cellular morphology, spatial organization, and population-level behavior~\cite{cell_segmentation_survey}. Across imaging modalities including brightfield, phase-contrast, and fluorescence, this task is challenging due to non-specific contrast between cell structures and background, high variability in cell morphology, noise, and unclear boundaries~\cite{multimodality_cell_segmentation_challenge,cell_segmentation_pmc,mais2020patchperpix}. In dense cultures and cytology specimens, cells frequently form overlapping clusters. Unlike natural-image occlusion, where the occluded region is often visually absent, many microscopy modalities exhibit semi-transparent overlap where the hidden cell structure remains partially visible but with weak contrast and mixed visual evidence~\cite{donet}. Overlapping regions, therefore, contain information from multiple instances, producing ambiguous boundaries and strong appearance entanglement between neighboring cells.




Resolving these overlapping instances requires amodal reasoning about the complete object extent from partial observations. Existing approaches address this through region-level decomposition~\cite{donet, gains, shapeformer, aisformer, vrspnet, orcnn, bcnet}, shape priors~\cite{c2fseg,shapemoe,vrspnet}, or multi-stage generative completion~\cite{mcdiffusion, sas}. While effective, these methods typically operate on local Region of Interest (RoI) features or depend on upstream predictions, limiting their ability to jointly reason over the full scene. For overlapping cells specifically, understanding how each instance relates to its neighbors is essential for determining which evidence belongs to which cell. Query-based segmentation models~\cite{mask2former, maskdino} provide a natural fit for this problem, as each instance is represented by a query that attends globally to image features and interacts with all other queries, enabling the inter-instance communication needed to reason about shared overlap regions in the full scene context.

In this work, we introduce \methodname{}, for overlapping cell instance segmentation. We argue that de-overlapping requires understanding the object structure in terms of its visible and hidden parts. To this end, we decompose each instance query into amodal, visible, and invisible sub-representations and recombine them with consistency regularization to improve complete object perception. 
To ensure that queries of overlapping cells remain distinguishable, we further introduce contrastive query learning that leverages denoising queries, which provide stable ground-truth-initialized representations during training, combining an instance-discriminative loss with a query alignment loss to separate queries in embedding space.

\vspace{0.18cm}
Our contributions are as follows:
\begin{itemize}[topsep=1.8pt,itemsep=0pt,parsep=0pt,label={--}]
    \item We propose query-level instance decomposition and recombination with consistency regularization to model amodal, visible, and invisible cell structure.
    \item We introduce DN-guided contrastive query learning with instance-discriminative and cosine alignment losses to improve separation of overlapping cells.
    \item We introduce a new Organoids benchmark for evaluating overlapping cell instance segmentation.\footnote{The Organoids dataset is available upon request.}
\end{itemize}

\section{Related Work}
\label{sec:related-work}

Segmenting overlapping instances requires reasoning about both object structure and inter-instance relationships, problems that have been explored from different angles in prior works.

\subsection{Amodal Instance Segmentation}
\label{sec:related-work-overlapping-amodal}

Instance segmentation has been built predominantly on two-stage detection frameworks. Mask R-CNN~\cite{maskrcnn} introduced a mask prediction head on top of Faster R-CNN \cite{faster_rcnn}, and subsequent methods such as Cascade R-CNN~\cite{cascadercnn} and HTC~\cite{htc} refined the multi-stage pipeline. 
These models serve as the foundation for overlapping and amodal segmentation methods.

\noindent\textbf{Direct methods.}
Occlusion-aware extensions of the two-stage pipeline introduce dedicated modules for handling overlapping instances. Occlusion R-CNN~\cite{occlusionrcnn} adds a bilayer decoupling head that separates occluder and occludee representations within each RoI to predict visible and amodal masks. BCNet~\cite{bcnet} predicts two overlapping layers via graph convolutional networks on RoI features. \cite{depthordering} formulates overlap as a depth-ordering problem, assigning layer indices to instances through a U-Net \cite{ronneberger2015unet} architecture. AISFormer~\cite{aisformer} brings transformer queries into the RoI pipeline, introducing mask tokens for occluder, visible, amodal, and invisible types that interact through self-attention. 
For cytology, DoNet~\cite{donet} introduces a decompose-and-recombine strategy that decomposes cell clusters into intersection and complement regions through a Dual-path Region Segmentation Module, followed by consistency-guided recombination. GAInS~\cite{gains} generates gradient anomaly maps that capture spatial regions of crossing, touching, and overlapping, and uses them to reweight the mask prediction loss in error-prone overlap regions.

\noindent\textbf{Shape-prior methods.}
Several approaches learn object shape distributions to complete occluded regions. C2F-Seg~\cite{c2fseg} and ShapeFormer~\cite{shapeformer} learn latent shape representations for coarse-to-fine amodal mask refinement. Prior-Guided Expansion~\cite{priorguidedexpansion} retrieves regression and flow transformations from a memory bank of shape priors. ShapeMoE~\cite{shapemoe} routes each instance to a specialized expert based on learned Gaussian shape embeddings. VRSP-Net~\cite{vrspnet} employs a diffusion-based shape prior estimation module conditioned on visible features. 
While effective in natural image domains with relatively consistent object geometries, these methods assume a learnable shape distribution that becomes problematic for cells exhibiting extreme morphological diversity.

\noindent\textbf{Generative methods.}
Diffusion-based approaches have recently been applied to amodal completion. pix2gestalt~\cite{pix2gestalt} uses conditional diffusion to synthesize complete objects from partial observations in a zero-shot manner. MC Diffusion~\cite{mcdiffusion} separates query objects from occluding context and applies progressive mixed-context diffusion for amodal completion. SAS~\cite{sas} formulates sequential amodal segmentation through cumulative occlusion learning, predicting amodal masks layer-by-layer from unoccluded to deeply occluded objects. These methods produce compelling completions but depend on upstream visible mask quality as conditioning input, creating pipeline dependencies where segmentation errors propagate into the completion stage.

\noindent\textbf{Foundation model adaptation.}
SAMBA~\cite{samba} proposes a SAM-based amodal segmentation foundation model with a separation-to-fusion structure for joint modal and amodal prediction. SAMEO~\cite{sameo} adapts the Segment Anything model for occluded scene understanding. These approaches must acquire overlap-handling behavior from data alone without explicit de-overlapping objectives, requiring large curated amodal datasets that are scarce in biomedical domains.


\subsection{DETR Models}
\label{sec:related-work-detr}

DETR~\cite{detr} reformulated object detection as a set prediction problem, where learnable queries attend to image features through a transformer decoder. Deformable DETR~\cite{deformabledetr} improved efficiency with multi-scale deformable attention. DN-DETR~\cite{dndetr} introduced denoising training by injecting noise-perturbed ground-truth labels as additional queries for reconstruction, improving convergence. DINO~\cite{dino} extended this idea with contrastive denoising groups and mixed query selection. Mask2Former~\cite{mask2former} introduced masked attention, restricting cross-attention to predicted foreground regions for segmentation. MaskDINO~\cite{maskdino} unified detection and segmentation by adding mask prediction through query-pixel dot products while inheriting denoising training. Unlike region-based methods that confine each instance to a local crop, these query-based architectures allow each query to attend globally over the full image, enabling inter-instance communication for reasoning about overlapping objects.

In the biomedical domain, IAUNet~\cite{iaunet} introduces a query-based U-Net architecture with a novel lightweight convolutional Pixel decoder and a Transformer decoder that refines object-specific features across multiple scales, demonstrating strong performance in cell segmentation. PCTrans~\cite{pctrans} uses position-guided cross-attention and contrastive losses on query embeddings to learn discriminative representations in dense biological scenes, though the method primarily targets crowded instances without explicit modeling of overlapping object structure. Despite their strong performance, mask transformers exhibit specific failure modes in dense scenes. DAC-DETR~\cite{dacdetr} shows that cross-attention gathers multiple queries toward the same object while self-attention disperses them to avoid duplicates, and learning these opposing effects jointly becomes increasingly difficult when nearby objects create conflicting signals. PanSR~\cite{pansr} demonstrates instance merging, where distinct objects collapse into one mask, and addresses it by constraining mask predictions with bounding box geometry.

\subsection{Contrastive Learning for Instance Discrimination}
\label{sec:related-work-contrastive}

In overlapping scenes, learning discriminative instance representations is essential for distinguishing objects that share similar appearance and spatial context. Contrastive learning provides a natural framework for shaping these representations.

\noindent\textbf{Category-level contrastive.}
Contrastive learning for DETR-based detectors has focused primarily on category-level query discrimination. CSPCL~\cite{cspcl} aligns content queries with category prototypes through intra-class attraction and inter-class repulsion losses, correcting missing semantic information for prohibited item detection in overlapping X-ray images. MMCL~\cite{mmcl} proposes a multi-class min-margin contrastive loss for anti-overlapping X-ray detection that balances intra-class diversity with inter-class separability. These methods improve category discrimination but do not address same-class instance separation, the primary challenge in cell segmentation, where all overlapping objects belong to the same category.

\noindent\textbf{Instance-level contrastive.}
Instance-level discriminative feature learning has been explored primarily in video instance segmentation, where temporal association provides natural positive and negative pairs. CAVIS~\cite{cavis} uses prototypical cross-frame contrastive loss to maintain instance embedding consistency across frames. VISAGE~\cite{visage} employs appearance-guided contrastive objectives for instance identity preservation across video frames. MDQE \cite{mdqe} mines discriminative query embeddings for video segmentation under occlusion through temporal cross-attention and inter-instance mask repulsion. ConQueR~\cite{conquer} and similar methods like \cite{ln-detr} embed ground-truth instances into the query space for contrastive training to reduce false positive predictions in 3D detection. These methods form contrastive pairs from temporal correspondences or ground-truth embeddings without imposing explicit geometric constraints on the pairwise similarity structure needed for de-overlapping. In our method, we combine discriminative feature learning with cosine alignment to ensure that queries of overlapping cells remain well-separated in the embedding space.

\section{Method}
\label{sec:method}

We address overlapping cell instance segmentation by extending MaskDINO with complementary objectives targeting object structure modeling and query discrimination in dense overlap scenes.

\begin{figure*}[t]
    \centering
    \includegraphics[width=\textwidth]{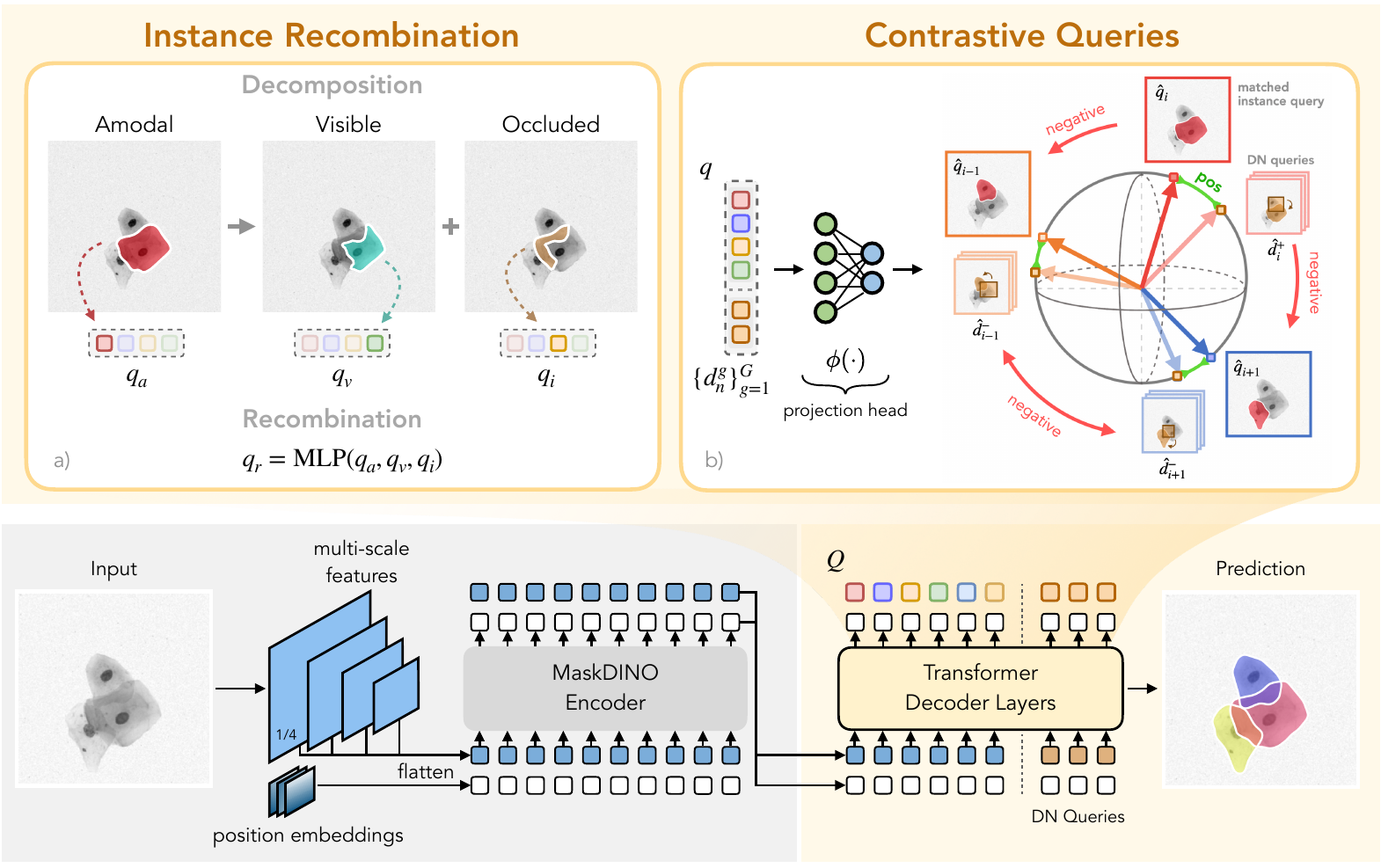}
    \caption{\textbf{Overview of QCell.}
QCell builds on a MaskDINO-style query-based segmentation pipeline, where multi-scale image features and positional embeddings are processed by the encoder and refined by transformer decoder layers with content and DN queries. The proposed modules are shown above: (a) instance recombination decomposes each query into amodal, visible, and occluded sub-representations and recombines them into a refined full-instance query; (b) contrastive query learning uses matched instance queries $\hat{q}_i$ by Hungarian matching as anchors, corresponding DN queries $\hat{d}^+_i$ across all groups as positives and other DN queries as negatives to align queries of the same cell and separate queries of different cells in latent space.}
    \label{fig:qcell_main}
\end{figure*}

\subsection{Preliminaries: MaskDINO}
\label{sec:preliminaries-maskdino}

Our method builds on MaskDINO, a unified query-based framework for detection and segmentation that extends DINO with a mask prediction branch.

\noindent\textbf{Architecture.}
Given an input image, a backbone network extracts multi-scale features, which are processed by a pixel decoder to produce multi-scale feature maps $\{F_l\}_{l=1}^{L}$, $F_l \in \mathbb{R}^{H_l \times W_l \times D}$, and a high-resolution pixel embedding map $F \in \mathbb{R}^{H \times W \times D}$. A set of $N$ learnable content query embeddings $Q \in \mathbb{R}^{N \times D}$ is iteratively refined through a stack of transformer decoder layers, each consisting of self-attention among queries, multi-scale deformable cross-attention with the feature maps, and a feed-forward network. After decoding, parallel prediction heads produce per-query classification scores $c_n \in \mathbb{R}^{K}$, bounding box coordinates $b_n \in \mathbb{R}^{4}$, and instance masks $M_n \in \mathbb{R}^{H \times W}$ obtained via dot product $M_n = q_n^\top F$. During training, the Hungarian algorithm assigns $M$ predictions to ground-truth instances through one-to-one bipartite matching, while the remaining $N - M$ queries are assigned to the \emph{"no object"} ($\varnothing$) class.


\noindent\textbf{DeNoising (DN) training.}
The decoder additionally receives DN queries constructed by adding random noise to the ground-truth bounding boxes and class labels, organized into $G$ denoising groups, each containing a noised version of all $M$ instances. We denote the DN query for instance $n$ in group $g$ as $d_n^g$. The decoder reconstructs clean targets from these noisy initializations, accelerating convergence. In our framework, we repurpose DN queries as stable anchors for contrastive learning (Section~\ref{sec:contrastive-query-learning}). The standard MaskDINO training objective is:

\begin{equation}
\mathcal{L}_{\mathrm{base}}
=
\lambda_{\mathrm{cls}}\mathcal{L}_{\mathrm{cls}}
+
\lambda_{\mathrm{box}}\mathcal{L}_{\mathrm{box}}
+
\lambda_{\mathrm{mask}}\mathcal{L}_{\mathrm{mask}}
+
\mathcal{L}_{\mathrm{dn}}.
\label{eq:base-loss}
\end{equation}

In overlapping cell segmentation, this formulation faces specific limitations. The single-mask prediction does not model the relationship between visible and occluded object parts, and the training objective lacks explicit supervision for learning discriminative instance features in dense overlapping scenes.

\subsection{Instance Recombination}
\label{sec:instance-recombination}

When semi-transparent cells overlap, the intersection region contains blended visual signals from both instances. Standard segmentation models produce a single mask per instance, which, due to limited perception capability in overlapping regions, makes it difficult to learn complete object structure when parts of the object are shared with or hidden by neighboring cells. Motivated by DoNet~\cite{donet}, we bring the decompose-and-recombine principle to the query level, removing the dependency on region proposals and enabling the model to reason about object structure through global attention.

\noindent\textbf{Decomposition.}
For each content query embedding $q$ in the decoder, we introduce three lightweight MLP heads that produce sub-query representations corresponding to the structural components of an instance:
\begin{equation}
q_k = \mathrm{MLP}_k(q),
\qquad
k \in \{a, v, i\},
\label{eq:ir-subquery-decomposition}
\end{equation}
where $q_a$, $q_v$, and $q_i$ encode the amodal (\textit{full extent}), visible (\textit{non-overlapped part}), and invisible (\textit{occluded part}) representations, respectively (see \cref{fig:qcell_main}). Each sub-query generates its corresponding mask through dot product with the pixel features $F$:
\begin{equation}
M_k = q_k^\top F,
\qquad
k \in \{a, v, i\}.
\label{eq:ir-submask-prediction}
\end{equation}

All queries in the decoder are passed through the decomposition heads. During training, matched queries are supervised with the corresponding ground-truth component masks. We supervise each sub-mask with binary cross-entropy and dice losses:

\begin{equation}
\mathcal{L}_{\mathrm{decomp}}
=
\sum_{k \in \{a, v, i\}}
\left[
\mathcal{L}_{\mathrm{bce}}(M_k, G_k)
+
\mathcal{L}_{\mathrm{dice}}(M_k, G_k)
\right],
\label{eq:ir-decomposition-loss}
\end{equation}
where $G_a$, $G_v$, $G_i$ are the ground-truth amodal, visible, and invisible masks.

\noindent\textbf{Recombination.}
After decomposition, we recombine the sub-query representations to produce a refined full-instance embedding. The three sub-queries are fused through a learned projection:
\begin{equation}
q_r = \mathrm{MLP}(q_a, q_v, q_i),
\label{eq:ir-recombination}
\end{equation}
which integrates information from all structural components into a single refined representation. The refined mask is then obtained as $M_r = q_r^\top F$ and supervised against the full amodal ground truth $G_a$:
\begin{equation}
\mathcal{L}_{\mathrm{refined}}
=
\mathcal{L}_{\mathrm{bce}}(M_r, G_a)
+
\mathcal{L}_{\mathrm{dice}}(M_r, G_a).
\label{eq:ir-refined-loss}
\end{equation}

The recombination step encourages the sub-queries to capture complementary information, since their fusion must recover the complete object. The refined query embedding $q_r$ encodes richer structural knowledge than the original query, having been trained to reason about both visible and hidden regions.

\noindent\textbf{Consistency regularization (CR).}
To enforce geometric coherence between the decomposed parts and the recombined prediction, we introduce a consistency regularization loss. The key constraint is that the refined mask should be recoverable from the union of the visible and invisible predictions:

\begin{equation}
\mathcal{L}_{\mathrm{CR}}
=
\mathcal{L}_{\mathrm{bce}}
\left(
M_r,\;
\mathtt{sg}
\left[
\mathrm{XOR}
\left(
\sigma(M_v) > 0.5,\;
\sigma(M_i) > 0.5
\right)
\right]
\right),
\label{eq:ir-consistency-loss}
\end{equation}

where $\sigma$ is the sigmoid function, $\mathtt{sg}(\cdot)$ denotes the stop-gradient operator, and the $\mathrm{XOR}(\cdot,\cdot)$ operation produces the recombined binary mask from the thresholded visible and invisible predictions. The binary mask is treated as a fixed pseudo-target, so gradients from $\mathcal{L}_{\mathrm{CR}}$ flow only through the refined prediction $M_r$.

The recombination loss is:
\begin{equation}
\mathcal{L}_{\mathrm{recomb}}
=
\mathcal{L}_{\mathrm{refined}}
+
\mathcal{L}_{\mathrm{CR}}.
\label{eq:ir-recombination-loss}
\end{equation}

The complete instance recombination loss is:
\begin{equation}
\mathcal{L}_{\mathrm{IR}}
=
\mathcal{L}_{\mathrm{decomp}}
+
\mathcal{L}_{\mathrm{recomb}}.
\label{eq:ir-loss}
\end{equation}

\subsection{Contrastive Query Learning}
\label{sec:contrastive-query-learning}

While the instance recombination module provides structural supervision for object decomposition, similar to other amodal methods~\cite{aisformer, donet, shapeformer}, it does not address representational similarity between queries of overlapping instances. Under heavy overlap, content query embeddings of nearby cells converge through self-attention, leading to representational collapse. Our key idea is decoupling the contrastive objective into two complementary requirements: (i) queries should capture distinctive instance features, and (ii) queries of co-occurring instances should remain sufficiently separated. We address both through contrastive losses on query embeddings within each decoder level.

\noindent\textbf{DeNoising queries as stable anchors.}
DN queries provide a natural foundation for contrastive learning. Since each ground-truth instance is guaranteed $G$ DN representations $\{d_n^g\}_{g=1}^{G}$ regardless of matching quality, they serve as reliable anchors. We empirically confirm their stability over matched queries in \cref{tab:dn-oracle}.

\noindent\textbf{Contrastive projection.}
We use a lightweight projection head $\phi$ that maps content query embeddings into a shared contrastive space (see \cref{fig:qcell_main}). Both DN queries and matched queries are projected through $\phi$, producing normalized embeddings. We denote the projected matched query for instance $n$ as $\hat{q}_n = \phi(q_n)$ and the projected DN query for instance $n$ in group $g$ as $\hat{d}_n^g = \phi(d_n^g)$. For each projected matched query $\hat{q}_n$, we define the positive set $\mathcal{P}_n = \{\hat{d}_n^g\}_{g=1}^{G}$ as DN embeddings for the same instance across all denoising groups, and the negative set $\mathcal{N}_n = \{\hat{d}_j^g \mid j \neq n,\; g \in \{1, \ldots, G\}\}$ as DN embeddings for all other instances.

\noindent\textbf{Instance-discriminative loss.}
To encourage the model to learn discriminative instance features, we formulate an InfoNCE-based objective. Using $\hat{q}_n$ as the anchor, $\mathcal{P}_n$ as positives, and $\mathcal{N}_n$ as negatives

\begin{equation}
\mathcal{L}_{\mathrm{disc}}
=
-
\sum_{n=1}^{M}
\log
\frac{
\sum\limits_{k^{+} \in \mathcal{P}_n}
\exp(\hat{q}_n \cdot k^{+} / \tau)
}{
\sum\limits_{k^{+} \in \mathcal{P}_n}
\exp(\hat{q}_n \cdot k^{+} / \tau)
+
\sum\limits_{k^{-} \in \mathcal{N}_n}
\exp(\hat{q}_n \cdot k^{-} / \tau)
}.
\label{eq:discriminative-loss}
\end{equation}

where $M$ is the number of matched instances and $\tau$ is a temperature parameter. This loss encourages each matched query to learn discriminative features that align with its corresponding DN representations while remaining distinct from DN representations of other instances.

\renewcommand{\arraystretch}{1.12}
\setlength{\tabcolsep}{2.2pt}
\begin{table*}[!t]
\centering
\tiny
\resizebox{\textwidth}{!}{
\begin{tabular}{l|c c c|c c c|c c c|c c c|c|c}
\multicolumn{1}{l}{} 
& \multicolumn{6}{c}{\textit{ISBI2014}} 
& \multicolumn{6}{c}{\textit{Revvity-25}}
& \multicolumn{2}{c}{} \\
\noalign{\hrule height 0.75pt}
Models\rule[-0.6ex]{0pt}{2.8ex} 
& AP & AP$_{50}$ & AP$_{75}$ & DICE & F1 & AJI
& AP & AP$_{50}$ & AP$_{75}$ & DICE & F1 & AJI
& \#params. & FLOPs \\
\hline
Mask R-CNN \cite{maskrcnn}
& 55.6 & 84.0 & 58.9 & 91.2 & 88.1 & 72.4
& 39.7 & 78.4 & 36.5 & 86.5 & 82.9 & 63.7
& 44M & 128G \\

ORCNN \cite{orcnn}
& 60.6 & 86.3 & 65.3 & 92.0 & 90.0 & 75.4
& 40.4 & 78.2 & 38.6 & 86.9 & 83.7 & 64.6
& 46M & 132G \\

VRSP-Net \cite{vrspnet}
& 51.1 & 81.6 & 53.4 & 91.6 & 88.5 & 73.1
& 39.1 & 77.7 & 34.7 & 87.7 & 85.4 & 67.2
& 99M & 118G \\

BCNet \cite{bcnet}
& 47.6 & 77.8 & 48.3 & 56.6 & 53.6 & 41.1
& 37.7 & 75.7 & 33.7 & 88.3 & 40.7 & 24.9
& 38M & 308G \\

AISFormer \cite{aisformer}
& 55.5 & 84.9 & 58.7 & 90.9 & 88.9 & 73.0
& 32.7 & 74.6 & 24.3 & 84.8 & 79.7 & 56.4
& 45M & 130G \\

PCTrans$^\dagger$ \cite{pctrans}
& 15.4 & 34.6 & 11.9 & 87.3 & 56.3 & 40.2
& 25.1 & 51.3 & 22.6 & 87.5 & 68.1 & 61.0
& 31M & 99G \\

GAInS \cite{gains}
& 54.8 & 83.9 & 59.0 & 90.9 & 89.0 & 73.0
& 38.4 & 77.0 & 34.8 & 86.6 & 82.2 & 63.6
& 53M & 117G \\

DoNet \cite{donet}
& 60.9 & 86.2 & \underline{66.1} & 92.1 & 89.6 & 75.2
& 44.6 & 81.5 & 45.1 & 87.6 & 85.4 & 67.5
& 73M & 117G \\

Mask2Former \cite{mask2former}
& 56.3 & 83.1 & 56.2 & 90.7 & 88.2 & 71.1
& 52.0 & 84.2 & \underline{58.5} & \underline{89.3} & 84.1 & 71.1
& 44M & 159G \\

MaskDINO \cite{maskdino}
& \underline{63.7} & \underline{89.0} & 65.4 & \underline{92.3} & \underline{90.0} & \underline{75.9}
& \underline{52.3} & \underline{85.5} & 57.5 & 89.3 & \textbf{87.1} & \underline{73.5}
& 44M & 163G \\

\rowcolor{our_results_color}
\textbf{QCell (ours)}
& \textbf{65.9} & \textbf{91.9} & \textbf{69.1} & \textbf{92.4} & \textbf{92.3} & \textbf{78.6}
& \textbf{52.9} & \textbf{86.0} & \textbf{59.3} & \textbf{89.4} & \underline{86.4} & \textbf{73.6}
& 45M & 182G \\
\hline 
\end{tabular}
}
\vspace{0.2cm}
\caption{\textbf{Comparison on ISBI2014 and Revvity-25.}
QCell achieves the best overall performance on both benchmarks, improving AP and AJI over prior amodal and cell instance segmentation methods. Rankings are determined from unrounded values. $^\dagger$ denotes our adapted implementation. Best results are shown in \textbf{bold}, and second-best results are \underline{underlined}.}
\label{tab:isbi-revvity-results}
\end{table*}

\noindent\textbf{Latent alignment loss.}
To impose direct geometric constraints on the embedding space, we complement the contrastive objective with a cosine alignment loss that explicitly controls the pairwise similarity structure. Using the same positive and negative sets $\mathcal{P}_n$ and $\mathcal{N}_n$

\begin{equation}
\mathcal{L}_{\mathrm{align}}
=
\sum_{n=1}^{M}
\left[
\sum_{k^{+} \in \mathcal{P}_n}
\left(
\cos(\hat{q}_n, k^{+}) - 1
\right)^2
+
\sum_{k^{-} \in \mathcal{N}_n}
\cos(\hat{q}_n, k^{-})^2
\right].
\label{eq:alignment-loss}
\end{equation}

The first term pulls matched predictions toward their corresponding DN representations across denoising groups, reinforcing identity consistency. The second term pushes matched predictions toward orthogonality with DN representations of other instances, directly penalizing the high cosine similarity that leads to representational collapse. Both losses are applied across multiple decoder layers.

\subsection{Training Objective}
\label{sec:training-objective}

The complete training objective combines the baseline MaskDINO losses with the three proposed components:
\begin{equation}
\mathcal{L}
=
\mathcal{L}_{\mathrm{base}}
+
\lambda_{\mathrm{IR}}\mathcal{L}_{\mathrm{IR}}
+
\lambda_{\mathrm{disc}}\mathcal{L}_{\mathrm{disc}}
+
\lambda_{\mathrm{align}}\mathcal{L}_{\mathrm{align}}.
\label{eq:total-loss}
\end{equation}

We compute losses on each decoder layer and sum them, following the auxiliary loss strategy in DETR-based architectures. Following \cite{maskdino}, we set $\lambda_{\text{cls}}=4.0$,
$\lambda_{\text{bce}}=5.0$, $\lambda_{\text{dice}}=5.0$,
$\lambda_{\text{box}}=5.0$, and $\lambda_{\text{giou}}=2.0$.
For the Instance Recombination module, the coarse mask BCE and Dice losses are
weighted by $5.0$, and the consistency loss by $1.0$. For the contrastive
objective, we set $\lambda_{\text{disc}}=2.0$ and
$\lambda_{\text{align}}=5.0$, with temperature $\tau=0.1$.

\renewcommand{\arraystretch}{1.12}
\setlength{\tabcolsep}{2.4pt}
\begin{table}[!t]
\centering
\tiny
\resizebox{0.78\columnwidth}{!}{
\begin{tabular}{l|c c c|c c c|c|c}
\multicolumn{9}{c}{\textit{Organoids}} \\
\noalign{\hrule height 0.75pt}
Models\rule[-0.6ex]{0pt}{2.8ex}
& AP & AP$_{50}$ & AP$_{75}$ & DICE & F1 & AJI & \#params. & FLOPs \\
\hline
Mask R-CNN \cite{maskrcnn}
& \underline{50.6} & 68.7 & \underline{55.9} & 93.9 & 68.6 & 61.7 & 44M & 128G \\
ORCNN \cite{orcnn}
& 48.7 & 67.1 & 54.3 & 93.4 & 70.6 & 62.4 & 46M & 132G \\
VRSP-Net \cite{vrspnet}
& 50.3 & 67.6 & 55.7 & 93.0 & \textbf{72.3} & 62.6 & 99M & 118G \\
BCNet \cite{bcnet}
& 50.3 & 67.7 & 55.3 & \underline{94.1} & 62.7 & 58.6 & 38M & 308G \\
AISFormer \cite{aisformer}
& 49.5 & 67.7 & 55.1 & 93.7 & 66.9 & 60.4 & 45M & 130G \\
PCTrans$^\dagger$ \cite{pctrans}
& 11.7 & 28.1 & 8.1 & 86.0 & 50.4 & 40.9 & 31M & 99G \\
GAInS \cite{gains}
& 48.8 & 67.1 & 54.4 & 93.5 & 66.6 & 59.9 & 53M & 117G \\
DoNet \cite{donet}
& 50.5 & 67.7 & 55.8 & \textbf{94.2} & 66.8 & 61.0 & 73M & 117G \\
Mask2Former$^\ddagger$ \cite{mask2former}
& 34.7 & 47.2 & 38.4 & 92.3 & 63.3 & 54.2 & 44M & 159G \\
Mask2Former \cite{mask2former}
& 31.3 & 46.9 & 33.9 & 91.0 & 48.2 & 36.2 & 44M & 185G \\
MaskDINO \cite{maskdino}
& 49.7 & \underline{68.7} & 54.7 & 92.2 & 71.5 & \underline{63.1} & 44M & 163G \\
\rowcolor{our_results_color}
\textbf{QCell (ours)}
& \textbf{51.0} & \textbf{69.1} & \textbf{56.4} & 92.5 & \underline{71.6} & \textbf{63.2} & 45M & 182G \\
\hline
\end{tabular}
}
\vspace{0.2cm}
\caption{\textbf{Comparison on Organoids.} $^\dagger$ denotes our adapted implementation of the method. $^\ddagger$ denotes a model trained with $N$=100 object queries. The best result is shown in \textbf{bold}, and the second-best result is \underline{underlined}.}
\label{tab:organoids-results}
\end{table}

\section{Experiments}
\label{sec:experiments}
In this section, we evaluate QCell on multiple datasets, including our novel Organoids benchmark for overlapping cell segmentation. We provide comprehensive comparisons with state-of-the-art methods and conduct ablation studies to demonstrate the effectiveness of each model component. We evaluate on three microscopy datasets that pose overlapping challenges and range in fine-grained details and object count across different imaging modalities:

\noindent\textbf{ISBI2014}~\cite{isbi2014} is a dataset from the Overlapping Cervical Cytology Image Segmentation Challenge. It includes 16 real extended depth-of-focus (EDF) cervical cytology images and 945 synthetic images with high-quality pixel-level annotations for nuclei and cytoplasm at a resolution of $512 \times 512$. We follow the challenge setting~\cite{isbi2014}, using 45 synthetic images for training, 90 for validation, and 810 for testing. Since our focus is on overlapping objects, we benchmark all models on cytoplasm annotations only, where the semi-transparent overlap between cells is most prevalent.

\noindent\textbf{Revvity-25}~\cite{iaunet} consists of 110 high-resolution $1080 \times 1080$ brightfield images, each containing on average 27 manually labeled and expert-validated cancer cells, totaling 2,937 annotated instances. The dataset provides highly accurate and detailed annotations for cell borders and overlap regions, making it a challenging benchmark for precise boundary delineation in dense scenes.

\begin{figure*}[t]
    \centering
    \includegraphics[width=\textwidth]{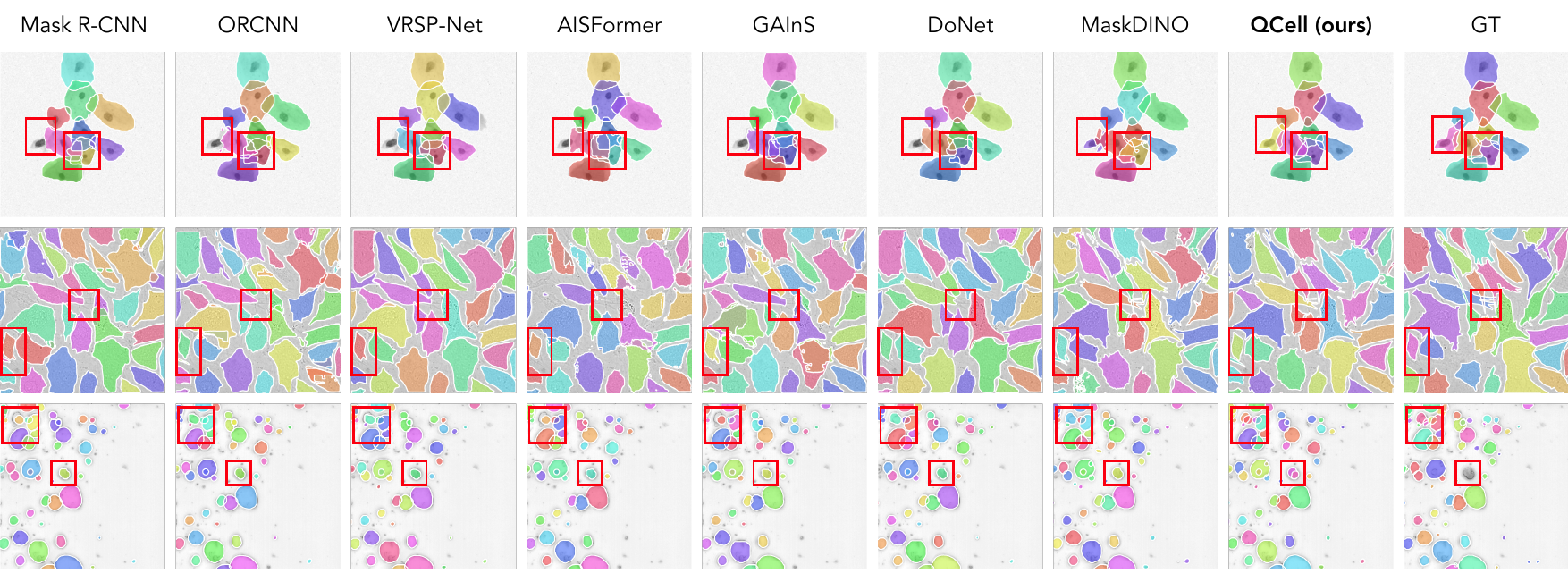}
    \vspace{-1.5em}
    \caption{\textbf{Qualitative comparison across datasets.}
    QCell produces more object-consistent instance masks across ISBI2014, Revvity-25, and Organoids. On ISBI2014, QCell better separates cells under heavy occlusion. On Revvity-25, it preserves fine cell details and boundaries. On Organoids, it maintains improved object consistency in dense overlapping scenes. Best viewed in color and zoomed in.}
    \label{fig:benchmark_predictions}
    \vspace{-1.0em}
\end{figure*}

\noindent\textbf{Organoids.} One of our key contributions is a novel Organoids dataset for overlapping object segmentation in brightfield microscopy. The dataset contains 1,186 training images, 1,199 validation images, and 201 test images at a resolution of $540 \times 540$. The dataset presents dense and highly overlapping scenes, with up to 105 instances per training image and an average of 96 instances per test image, reaching a maximum of 223. This makes Organoids a challenging real-world benchmark for evaluating instance separation and de-overlapping in microscopy. Additional dataset details are provided in the \textcolor{suppred}{Supplementary Material}.
\vspace{-0.5em}

\subsection{Implementation Details}
\label{sec:implementation_details}
We adopt Detectron2-based model implementations for all experiments and use a ResNet-50-FPN backbone with ImageNet-pretrained weights. All reported models, including MaskDINO, are trained to predict full amodal masks. Following \cite{donet}, during training, we adopt SGD with momentum $0.9$ as the optimizer for R-CNN-based models. We set the initial learning rate to $10^{-3}$ and use linear warm-up during the first 1k iterations. For query-based models, we use a consistent training schedule across all benchmarks. All query-based models use 100 queries on ISBI2014 and Revvity-25 and 300 on Organoids. We train all query-based models with AdamW \cite{adamw}, using an initial learning rate of $10^{-4}$, weight decay $0.05$, and a backbone learning-rate multiplier of $0.1$. All models are trained on ISBI2014, Revvity-25, and Organoids for 60k iterations with a batch size of 2, decreasing the learning rate by a factor of $0.1$ at 50k and 55k iterations. During training, we select the best checkpoint based on validation-set performance and report test results averaged over three random seeds. Following \cite{donet}, we use the same evaluation protocol for all methods. All experiments are conducted on a single NVIDIA H200 Tensor Core GPU with 141 GB of HBM3e memory.
\vspace{-0.5em}

\subsection{Main Results}
\label{sec:main_results}
We compare QCell with state-of-the-art amodal and cell instance segmentation methods on ISBI2014, Revvity-25, and Organoids. As shown in \cref{tab:isbi-revvity-results}, QCell achieves the best performance on ISBI2014 cytoplasm segmentation, outperforming all compared methods with $65.9$ AP, $92.3$ F1, and $78.6$ AJI. Compared to the MaskDINO baseline, QCell improves AP by $+2.2$, F1 by $+2.3$, and AJI by $+2.7$. We observe that QCell produces more object-consistent masks compared to other models, especially in regions where neighboring cells share weak or semi-transparent boundaries. Two-stage R-CNN-based methods are limited by detection quality and often struggle to segment highly overlapping instances, while QCell better preserves separated object masks in these cases, as shown in \cref{fig:benchmark_predictions}. On Revvity-25, QCell also achieves the best AP and AJI among the compared methods, reaching $52.9$ AP, $86.4$ F1, and $73.6$ AJI. The model produces visually cleaner segmentations with fewer false positives in dense regions. As shown in \cref{fig:benchmark_predictions}, query-based models preserve fine cell structures better than R-CNN-based methods, which often miss thin boundaries or merge nearby instances under strong overlap. On Organoids, QCell remains competitive in the most densely populated benchmark while achieving the best overall results, with $51.0$ AP, $71.6$ F1, and $63.2$ AJI. The dataset contains substantially more instances per image than ISBI2014 and Revvity-25, making accurate instance ranking and separation more difficult. Despite this, QCell achieves the best AP and AJI among the compared methods, showing that the proposed query-based de-overlapping strategy generalizes to high-density microscopy scenes. In \cref{tab:organoids-results}, we report Mask2Former trained with both $N=100$ and $N=300$ object queries on Organoids. The $N=300$ configuration matches the query count used by the other query-based models, while the $N=100$ configuration is retained because increasing the number of queries degraded Mask2Former performance in our experiments. We also note that PCTrans differs from the other instance segmentation methods in its mask representation. The framework predicts categorical masks, where each pixel receives a single label. We train and report PCTrans using the provided setup and observe limited generalization to overlapping amodal cell segmentation. The model obtains reasonable F1 and AJI scores. The AP stays low as the model doesn't produce per-instance confidence scores, thus the AP has no meaningful ranking.

\renewcommand{\arraystretch}{1.15}
\setlength{\tabcolsep}{2.6pt}
\begin{table*}[!t]
\centering
\tiny
\resizebox{\textwidth}{!}{
\begin{tabular}{c c c c c|c c c|c c c}
\noalign{\hrule height 0.75pt}
$\mathcal{L}_{\mathrm{base}}$
& $\mathcal{L}_{\mathrm{decomp}}$
& $\mathcal{L}_{\mathrm{recomb}}$
& $\mathcal{L}_{\mathrm{disc}}$
& $\mathcal{L}_{\mathrm{align}}$
& AP & AP$_{50}$ & AP$_{75}$ & DICE & F1 & AJI \\
\hline
\checkmark &  &  &  &
& 63.7 & 89.0 & 65.4 & 92.3 & 90.0 & 75.9 \\
\hline
\checkmark & \checkmark &  &  &
& 65.1 & 89.8 & 68.1 & 92.6 & 90.0 & 76.2 \\
\checkmark & \checkmark & \checkmark &  &
& \underline{66.6} & 90.1 & \textbf{69.2} & \textbf{92.8} & 90.7 & 77.1 \\
\hline
\checkmark &  &  & \checkmark &
& 65.5 & 89.8 & 67.5 & 92.6 & 90.6 & 76.7 \\
\checkmark &  &  &  & \checkmark
& 66.2 & 89.7 & 68.6 & 92.6 & 90.5 & 76.4 \\
\checkmark &  &  & \checkmark & \checkmark
& \textbf{67.0} & \underline{91.0} & \underline{69.2}
& \underline{92.7} & \underline{91.6} & \underline{77.9} \\
\hline
\rowcolor{our_results_color}
\checkmark & \checkmark & \checkmark & \checkmark & \checkmark
& 65.9 & \textbf{91.9} & 69.1 & 92.4 & \textbf{92.3} & \textbf{78.6} \\
\hline
\end{tabular}
}
\vspace{-0.5em}
\caption{Ablation study on ISBI2014. $\mathcal{L}_{\mathrm{base}}$ denotes the standard MaskDINO objective, $\mathcal{L}_{\mathrm{decomp}}$ the decomposition loss, $\mathcal{L}_{\mathrm{recomb}}=\mathcal{L}_{\mathrm{refined}}+\mathcal{L}_{\mathrm{CR}}$ the recombination loss, $\mathcal{L}_{\mathrm{disc}}$ the instance-discriminative loss, and $\mathcal{L}_{\mathrm{align}}$ the cosine alignment loss.}
\label{tab:ablation-study}
\vspace{-2.5em}
\end{table*}

\renewcommand{\arraystretch}{1.15}
\setlength{\tabcolsep}{2.6pt}
\begin{table}[!t]
\centering
\tiny
\resizebox{0.8\columnwidth}{!}{
\begin{tabular}{c c c c c|c c c}
\noalign{\hrule height 0.75pt}
$\mathcal{L}_{\mathrm{base}}$
& $\mathcal{L}_{\mathrm{decomp}}$
& $\mathcal{L}_{\mathrm{recomb}}$
& $\mathcal{L}_{\mathrm{disc}}$
& $\mathcal{L}_{\mathrm{align}}$
& AP & AP$_{50}$ & AP$_{75}$ \\
\hline
\checkmark &  &  &  & 
& 11.67 & 29.09 & 7.64 \\
\hline
\checkmark & \checkmark &  &  & 
& 12.12 & 28.71 & \underline{8.93} \\
\checkmark & \checkmark & \checkmark &  & 
& 12.29 & 29.52 & 8.92 \\
\hline
\checkmark &  &  & \checkmark & \checkmark
& \underline{12.63} & \textbf{30.21} & 8.50 \\
\hline
\rowcolor{our_results_color}
\checkmark & \checkmark & \checkmark & \checkmark & \checkmark
& \textbf{13.68} & \underline{30.17} & \textbf{10.97} \\
\hline
\end{tabular}
}
\vspace{0.2cm}
\caption{Ablation results on highly overlapping ISBI2014 instances. The subset includes instances whose ground-truth mask has pairwise IoU $\geq 0.5$ with another instance. Combining Instance Recombination and contrastive query learning gives the strongest performance under severe overlap.}
\label{tab:high-overlap}
\vspace{-0.5em}
\end{table}

\subsection{Ablation Studies}
\label{sec:ablation_studies}

In this section, we ablate the main components of \methodname{} on ISBI2014.

\noindent\textbf{Instance Recombination.}
The Instance Recombination (IR) module improves complete object reasoning under overlap by decomposing each query into amodal, visible, and invisible sub-representations and recombining them into a refined query. As shown in \cref{tab:ablation-study}, adding the instance recombination loss improves the MaskDINO baseline from $63.7$ to $65.1$ AP and from $65.4$ to $68.1$ AP$_{75}$. Adding Consistency Regularization (CR) further improves performance, reaching $66.6$ AP, $69.2$ AP$_{75}$, and $77.1$ AJI. Enforcing consistency between the visible, invisible, and recombined full masks helps produce more coherent object predictions in overlapping regions.


\noindent\textbf{Contrastive Query Learning.}
In overlapping regions, object features from neighboring cells mix, making it difficult for queries to learn distinctive instance representations. To investigate the potential of improved query representations, we conduct a DN oracle analysis (\cref{tab:dn-oracle}), where main-query predictions are matched to ground-truth instances using the Hungarian matcher and replaced with their corresponding DN-query predictions at test time. We report the mean performance over all DN queries across three random seeds. This improves performance from $63.7$ to $68.6$ AP and from $75.9$ to $80.0$ AJI, suggesting that DN queries capture higher-quality instance representations and motivating their use as supervision anchors for contrastive query learning. \cref{tab:ablation-study} shows the effect of each contrastive loss. Adding $\mathcal{L}_{\mathrm{disc}}$ alone improves the baseline to $65.5$ AP, while $\mathcal{L}_{\mathrm{align}}$ alone reaches $66.2$ AP, indicating that explicit latent separation provides a stronger individual signal. Combining both losses yields $67.0$ AP and $77.9$ AJI, demonstrating that discriminative feature learning and latent alignment are complementary. In the full model, recombination also restores amodal masks for heavily occluded cells missed by the baseline, improving recall and AP$_{50}$ while reducing FNo. Since their hidden regions must be inferred, the reconstructed masks can have moderate IoU, which can limit AP$_{75}$ and threshold-averaged AP despite detecting more cells overall. The full \methodname{} model with all components achieves the best F1 ($92.3$) and AJI ($78.6$), with AP$_{50}$ reaching $91.9$.

\renewcommand{\arraystretch}{1.15}
\setlength{\tabcolsep}{1.5pt}
\begin{table}[b]
\centering
\tiny
\resizebox{0.8\columnwidth}{!}{
\begin{tabular}{l|c c c|c c c}
\noalign{\hrule height 0.75pt}
Models\rule[-0.6ex]{0pt}{2.8ex}
& AP & AP$_{50}$ & AP$_{75}$ & DICE & F1 & AJI \\
\hline
Baseline
& 63.7 & 89.0 & 65.4 & 92.3 & 90.0 & 75.9 \\
\textbf{DN oracle}
& $\mathbf{68.6}$\smallstd{1.2}
& $\mathbf{93.5}$\smallstd{0.6}
& $\mathbf{71.9}$\smallstd{1.7}
& $\mathbf{92.4}$\smallstd{0.2}
& $\mathbf{94.1}$\smallstd{0.3}
& $\mathbf{80.0}$\smallstd{0.6} \\
\hline
\end{tabular}
}
\vspace{0.2cm}
\caption{DN oracle analysis on ISBI2014. Replacing main-query predictions matched to ground-truth instances using the Hungarian matcher with their corresponding DN-query predictions improves all metrics, showing that DN queries often provide cleaner instance representations and motivating their use for contrastive query learning.}
\label{tab:dn-oracle}
\end{table}

\noindent\textbf{Performance under Heavy Overlap.}
We further analyze performance on the high-overlap subset, containing instances whose ground-truth mask has pairwise IoU $\geq 0.50$ with another instance. \cref{tab:high-overlap} show that under severe overlap, the baseline achieves only $11.67$ AP, reflecting the difficulty of separating instances that share a large portion of their spatial extent. Adding instance recombination with consistency regularization improves AP from $11.67$ to $12.29$ by providing structural supervision for visible and hidden object parts. Contrastive query learning reaches $12.63$ AP, with AP$_{50}$ improving from $29.09$ to $30.21$, indicating that improved instance discrimination benefits heavily overlapping cases. Combining both components yields the strongest result at $13.68$ AP and $10.97$ AP$_{75}$, a $+2.01$ AP and $+3.33$ AP$_{75}$ gain over the baseline. This confirms that instance recombination and contrastive query alignment address complementary aspects of the de-overlapping problem.

\noindent\textbf{Object-level Error Analysis.}
To distinguish missed or merged instances from pixel-level coverage, \cref{tab:object-level-analysis} reports the object-based false-negative rate (FNo) and pixel-based true-positive rate (TPp). QCell achieves the lowest FNo and highest TPp across all three datasets, indicating improved instance recovery and separation while maintaining strong pixel coverage. Relative to MaskDINO, DICE remains comparable across the three benchmarks (\cref{tab:isbi-revvity-results,tab:organoids-results}), suggesting that the gains primarily arise from improved de-overlapping rather than boundary refinement. QCell achieves these improvements with 45M parameters and 182G FLOPs, a modest increase over MaskDINO with 44M parameters and 163G FLOPs.

\renewcommand{\arraystretch}{1.15}
\setlength{\tabcolsep}{1.5pt}
\begin{table}[!t]
\setlength{\tabcolsep}{2.6pt}
\centering
\small
\resizebox{\textwidth}{!}{
\begin{tabular}{l|c c|c c|c c|c|c}
\multicolumn{1}{l}{}
& \multicolumn{2}{c}{\textit{ISBI2014}}
& \multicolumn{2}{c}{\textit{Organoids}}
& \multicolumn{2}{c}{\textit{Revvity-25}}
& \multicolumn{2}{c}{} \\
\noalign{\hrule height 0.75pt}
Models\rule[-0.6ex]{0pt}{2.8ex}
& FNo $\downarrow$ & TPp $\uparrow$
& FNo $\downarrow$ & TPp $\uparrow$
& FNo $\downarrow$ & TPp $\uparrow$
& \#params. & FLOPs \\
\hline
Mask R-CNN \cite{maskrcnn}
& $17.3$\smallstd{0.7} & $95.6$\smallstd{0.1}
& $43.1$\smallstd{0.7} & $93.4$\smallstd{0.1}
& $17.7$\smallstd{0.2} & $85.0$\smallstd{0.9}
& 44M & 128G \\

ORCNN \cite{orcnn}
& $14.8$\smallstd{0.9} & $95.1$\smallstd{0.3}
& $40.7$\smallstd{0.9} & $93.3$\smallstd{0.0}
& $17.5$\smallstd{0.0} & $83.4$\smallstd{0.0}
& 46M & 132G \\

VRSP-Net \cite{vrspnet}
& $15.3$\smallstd{1.0} & $93.8$\smallstd{0.1}
& $32.2$\smallstd{2.2} & $93.6$\smallstd{0.3}
& $16.1$\smallstd{0.2} & $86.0$\smallstd{0.4}
& 99M & 118G \\

BCNet \cite{bcnet}
& $64.5$\smallstd{7.6} & $45.7$\smallstd{19.0}
& $50.3$\smallstd{0.0} & $94.2$\smallstd{0.0}
& $73.4$\smallstd{0.1} & $86.9$\smallstd{0.3}
& 38M & 308G \\

AISFormer \cite{aisformer}
& $16.2$\smallstd{1.5} & $94.9$\smallstd{0.2}
& $44.4$\smallstd{0.7} & $93.1$\smallstd{0.4}
& $20.7$\smallstd{2.2} & $83.1$\smallstd{1.5}
& 45M & 130G \\

PCTrans \cite{pctrans}
& $55.7$\smallstd{11.5} & $92.2$\smallstd{0.4}
& $46.6$\smallstd{9.0} & $88.2$\smallstd{0.0}
& $18.5$\smallstd{1.8} & $87.0$\smallstd{0.4}
& 31M & 99G \\

GAInS \cite{gains}
& $17.3$\smallstd{2.9} & $95.8$\smallstd{0.1}
& $43.9$\smallstd{2.0} & $93.3$\smallstd{0.4}
& $18.7$\smallstd{0.2} & $84.8$\smallstd{0.1}
& 53M & 117G \\

DoNet \cite{donet}
& $15.4$\smallstd{1.1} & $95.2$\smallstd{0.5}
& $44.0$\smallstd{1.5} & $\underline{94.4}$\smallstd{0.3}
& $15.2$\smallstd{0.2} & $86.7$\smallstd{0.3}
& 73M & 117G \\

Mask2Former \cite{mask2former}
& $13.3$\smallstd{1.5} & $96.3$\smallstd{0.4}
& $32.6$\smallstd{1.5} & $92.5$\smallstd{0.8}
& $12.1$\smallstd{0.2} & $\underline{89.1}$\smallstd{1.0}
& 44M & 159G \\

MaskDINO \cite{maskdino}
& $\underline{11.6}$\smallstd{2.7} & $\underline{96.2}$\smallstd{0.2}
& $\underline{30.1}$\smallstd{4.7} & $93.7$\smallstd{0.7}
& $\underline{11.9}$\smallstd{0.1} & $88.4$\smallstd{0.1}
& 44M & 163G \\

\rowcolor{our_results_color}
\textbf{QCell (ours)}
& $\mathbf{8.7}$\smallstd{0.5} & $\mathbf{96.9}$\smallstd{0.2}
& $\mathbf{27.5}$\smallstd{0.1} & $\mathbf{94.5}$\smallstd{0.1}
& $\mathbf{11.7}$\smallstd{0.1} & $\mathbf{89.4}$\smallstd{0.1}
& 45M & 182G \\
\hline
\end{tabular}
}
\caption{\textbf{Segmentation quality and model efficiency.} Object-based false-negative rate (FNo) and pixel-based true-positive rate (TPp) across the three datasets, with parameter counts and FLOPs. FLOPs are averaged over the ISBI2014 test set.}
\label{tab:object-level-analysis}
\vspace{-0.5em}
\end{table}

\vspace{-0.5em}
\section{Conclusions}
\label{sec:conclusions}
We presented QCell, a query-based model for de-overlapping cell instance segmentation in microscopy. Our approach introduces two complementary components that address the structural and representational challenges of overlapping semi-transparent cells. The instance recombination module decomposes and recombines query representations in latent space, enabling the model to reason about complete object structure with consistency regularization. The contrastive query alignment objective leverages denoising queries as stable per-instance anchors and combines an instance-discriminative loss with a cosine alignment loss to learn distinctive and well-separated instance features. We additionally introduced a new Organoid dataset benchmark for overlapping cell segmentation in dense brightfield microscopy scenes. Experiments on ISBI2014, Revvity-25, and Organoids demonstrate that QCell significantly outperforms competing methods, achieving state-of-the-art performance in overlapping cell segmentation.

%
\section*{Acknowledgments}
The authors acknowledge the support of the European Union and the Estonian Research Council through project TEM-TA101. Computational resources were provided by the High-Performance Computing Cluster at the University of Tartu. We thank the Biomedical Computer Vision Lab for its invaluable support. We thank Revvity and the \textit{Institut de Recherche en Santé Digestive} (IRSD), Inserm UMR 1220, Toulouse, France, for jointly providing the Organoids dataset and supporting its annotation and validation. We express our gratitude to the Armed Forces of Ukraine and the bravery of the Ukrainian people for enabling a secure working environment, without which this work would not have been possible.


\bibliography{egbib}

\end{document}